\documentclass[10pt,twocolumn,letterpaper]{article}

\usepackage{iccv}
\usepackage{times}
\usepackage{epsfig}
\usepackage{graphicx}
\usepackage{amsmath}
\usepackage{amssymb}

\usepackage{color}
\usepackage{booktabs}
\usepackage{multirow}
\usepackage{array}
\usepackage{xcolor}
\usepackage{subcaption}
\usepackage{graphicx}
\usepackage{hhline}
\usepackage{makecell}
\usepackage{colortbl}
\usepackage[normalem]{ulem}
\usepackage{soul}
\usepackage{url}
\definecolor{gray}{rgb}{0.7,0.7,0.7}
\definecolor{darkred}{rgb}{0.75,0.0,0.0}

\usepackage[square,comma,numbers]{natbib}

\usepackage[pagebackref=true,breaklinks=true,letterpaper=true,colorlinks,bookmarks=false]{hyperref}

\usepackage[british,UKenglish,USenglish,english,american]{babel}
\newcommand{\br}[2]{$\langle#1, #2\rangle$}

\newcolumntype{x}{>\small c}
\newcolumntype{L}[1]{>{\raggedright\let\newline\\\arraybackslash\hspace{0pt}}m{#1}}
\newcolumntype{C}[1]{>{\centering\let\newline\\\arraybackslash\hspace{0pt}}m{#1}}
\newcolumntype{R}[1]{>{\raggedleft\let\newline\\\arraybackslash\hspace{0pt}}m{#1}}

\newcolumntype{o}{>\small L}

\usepackage[pagebackref=true,breaklinks=true,letterpaper=true,colorlinks,bookmarks=false]{hyperref}

\iccvfinalcopy 

\def\iccvPaperID{***} 

\ificcvfinal\fi
\begin{document}

\title{Transitive Invariance for Self-supervised Visual Representation Learning}

\author{Xiaolong Wang$^{1}$  \quad     \quad  Kaiming He$^{2}$    \quad  \quad  Abhinav Gupta$^{1}$  \\
$^1$Carnegie Mellon University \quad $^2$Facebook AI Research \\
}
\maketitle

\begin{abstract}
Learning visual representations with self-supervised learning has become popular in computer vision. The idea is to design auxiliary tasks where labels are free to obtain. Most of these tasks end up providing data to learn specific kinds of invariance useful for recognition. In this paper, we propose to exploit different self-supervised approaches to learn representations invariant to (i) inter-instance variations (two objects in the same class should have similar features) and (ii) 
intra-instance variations (viewpoint, pose, deformations, illumination, \etc). Instead of combining two approaches with multi-task learning, we argue to organize and reason the data with multiple variations. Specifically, we propose to generate a graph with millions of objects mined from hundreds of thousands of videos. The objects are connected by two types of edges which correspond to two types of invariance: ``different instances but a similar viewpoint and category'' and ``different viewpoints of the same instance''. By applying simple transitivity on the graph with these edges, we can obtain pairs of images exhibiting richer visual invariance.
We use this data to train a Triplet-Siamese network with VGG16 as the base architecture and apply the learned representations to different recognition tasks. For object detection, we achieve $63.2\%$ mAP on PASCAL VOC 2007 using Fast R-CNN (compare to $67.3\%$ with ImageNet pre-training). For the challenging COCO dataset, our method is surprisingly close ($23.5\%$) to the ImageNet-supervised counterpart ($24.4\%$) using the Faster R-CNN framework. We also show that our network can perform significantly better than the ImageNet network in the surface normal estimation task. 
\end{abstract}

\section{Introduction}

\begin{figure}
\centering
\includegraphics[width=0.95\linewidth]{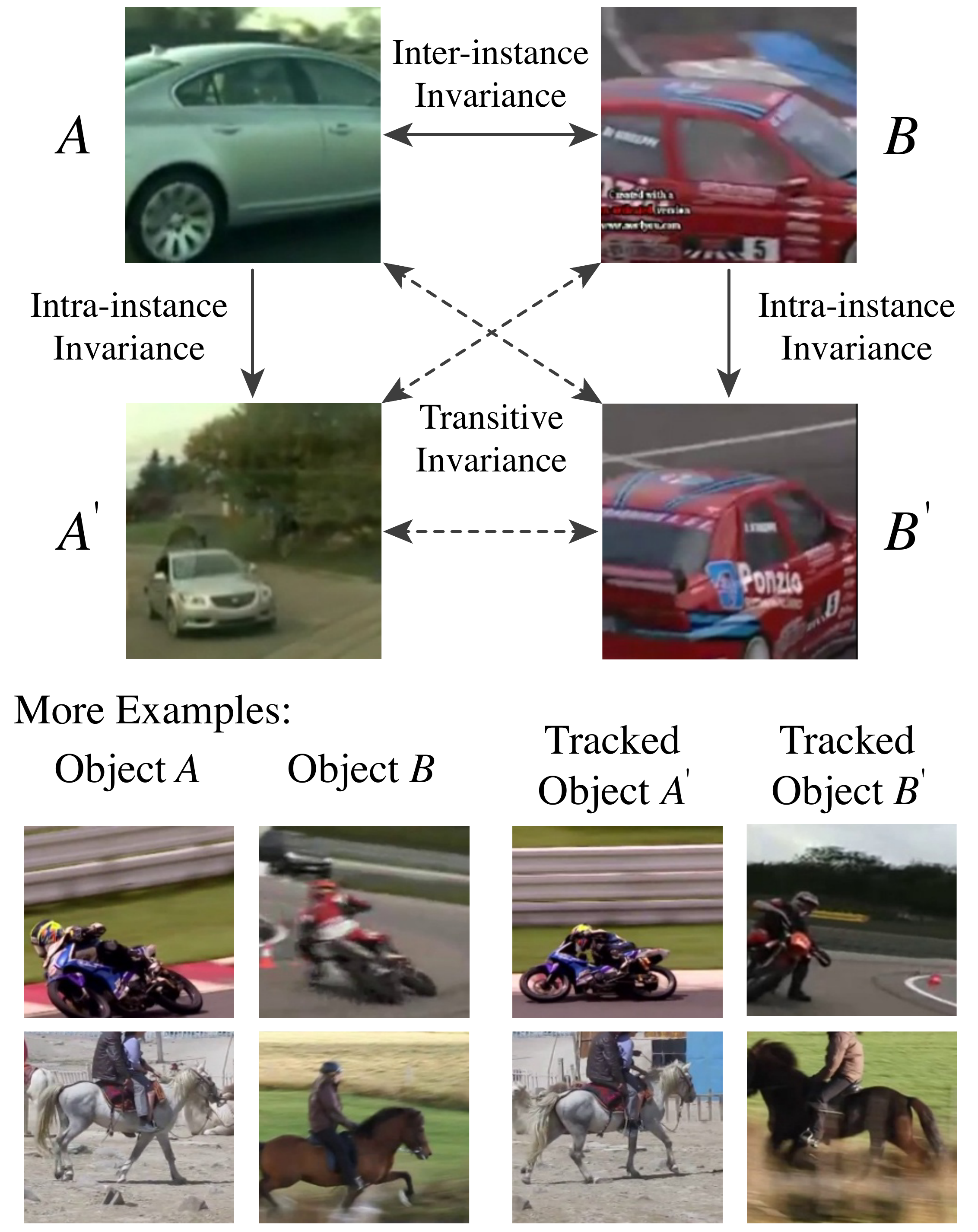}
\caption{We propose to obtain rich invariance by applying simple transitive relations. In this example, two different cars $A$ and $B$ are linked by the features that are good for inter-instance invariance (\eg, using \cite{Doersch2015}); and each car is linked to another view ($A'$ and $B'$) by visual tracking \cite{Wang2015}.
Then we can obtain new invariance from object pairs \br{A}{B'}, \br{A'}{B}, and \br{A'}{B'} via transitivity. We show more examples in the bottom.}
\label{fig:teaser}
\end{figure}

Visual invariance is a core issue in learning visual representations. Traditional features like SIFT \cite{Lowe2004} and HOG \cite{Dalal2005} are histograms of edges that are to an extent invariant to illumination, orientations, scales, and translations. Modern deep representations are capable of learning high-level invariance from large-scale data \cite{Olga15} , \eg, viewpoint, pose, deformation, and semantics. These can also be transferred to complicated visual recognition tasks \cite{Girshick2014,Long2015}.

In the scheme of supervised learning, human annotations that map a variety of examples into a single label provide supervision for learning invariant representations. For example, two horses with different illumination, poses, and breeds are invariantly annotated as a category of ``horse''. Such human knowledge on invariance is expected to be learned by capable deep neural networks \cite{LeCun1989,Krizhevsky2012} through carefully annotated data. However, large-scale, high-quality annotations come at a cost of expensive human effort.

Unsupervised or ``self-supervised'' learning (\eg, \cite{Wang2015,Doersch2015,Pinto2016,Zhang2016,Zhang2017,Li2016,Pathak2017,Amir16,Luo2017,Zhou2017}) recently has attracted increasing interests because the ``labels'' are free to obtain. Unlike supervised learning that learns  invariance from the semantic labels, the self-supervised learning scheme mines it from the nature of the data. We observe that most self-supervised approaches learn representations that are invariant to: (i) \emph{inter-instance} variations, which reflects the commonality among different instances. For example, relative positions of patches \cite{Doersch2015} (see also Figure~\ref{fig:context}) or channels of colors \cite{Zhang2016,Zhang2017} can be predicted through the commonality shared by many object instances; (ii) \emph{intra-instance} variations. Intra-instance invariance is learned from the pose, viewpoint, and illumination changes by tracking a single moving instance in videos \cite{Wang2015,Pathak2017}. However, either source of invariance can be as rich as that provided by human annotations on large-scale datasets like ImageNet.

Even after significant advances in the field of self-supervised learning, there is still a long way to go compared to supervised learning. What should be the next steps? It seems that an obvious way is to obtain multiple sources of invariance by combining multiple self-supervised tasks, \eg, via multiple losses. Unfortunately, this na\"ive solution turns out to give little improvement (as we will show by experiments).

We argue that the trick lies not in the tasks but in the way of exploiting data. To leverage both intra-instance and inter-instance invariance, in this paper we construct a huge affinity graph consisting of two types of edges (see Figure~\ref{fig:teaser}): the first type of edges relates ``different instances of similar viewpoints/poses and potentially the same category'', and the second type of edges relates ``different viewpoints/poses of an identical instance''. We instantiate the first type of edges by learning commonalities across instances via the approach of \cite{Doersch2015}, and the second type by unsupervised tracking of objects in videos \cite{Wang2015}. We set up simple transitive relations on this graph to infer more complex invariance from the data, which are then used to train a Triplet-Siamese network for learning visual representations. 

Experiments show that our representations learned without any annotations can be well transferred to the object detection task. Specifically, we achieve $63.2\%$ mAP with VGG16 \cite{Simonyan2014} when fine-tuning Fast R-CNN on VOC2007, against the ImageNet pre-training baseline of $67.3\%$.
More importantly, we also report the first-ever result of un-/self-supervised pre-training models fine-tuned on the challenging COCO object detection dataset \cite{Lin2014}, achieving $23.5\%$ AP comparing against $24.4\%$ AP that is fine-tuned from an ImageNet pre-trained counterpart (both using VGG16).
To our knowledge, this is the closest accuracy to the ImageNet pre-training counterpart obtained on object detection tasks.


\section{Related Work} 
Unsupervised learning of visual representations is a research area of particular interest. Approaches to unsupervised learning can be roughly categorized into two main streams: (i) generative models, and (ii) self-supervised learning. Earlier methods for generative models include Anto-Encoders \cite{Olshausen1997,Vincent2008,Lee2009,Le2012} and Restricted Boltzmann Machines (RBMs) \cite{Hinton2006,Bengio2007,Tang2012, Eslami2012}. For example, Le \etal~\cite{Le2012} trained a multi-layer auto-encoder on a large-scale dataset of YouTube videos: although no label is provided, some neurons in high-level layers can recognize cats and human faces.
Recent generative models such as Generative Adversarial Networks \cite{Goodfellow2014} and Variational Auto-Encoders \cite{Kingma2014} are capable of generating more realistic images. The generated examples or the neural networks that learn to generate examples can be exploited to learn representations of data \cite{Dumoulin2016,Donahue2016}.

Self-supervised learning is another popular stream for learning invariant features.
Visual invariance can be captured by the same instance/scene taken in a sequence of video frames \cite{Wang2015,Srivastava2015,Jayaraman2015,Agrawal2015,Mathieu2015,Walker2016,Li2016,Pathak2017,misra2016unsupervised,Goroshin2015}. For example, Wang and Gupta \cite{Wang2015} leverage tracking of objects in videos to learn visual invariance within individual objects; Jayaraman and Grauman \cite{Jayaraman2015} train a Siamese network to model the ego-motion between two frames in a scene; Mathieu \etal \cite{Mathieu2015} propose to learn representations by predicting future frames; Pathak \etal \cite{Pathak2017} train a network to segment the foreground objects where are acquired via motion cues. On the other hand, common characteristics of different object instances can also be mined from data \cite{Doersch2015,Zhang2016,Zhang2017,larsson2016learning,larsson2017colorproxy}. For example, relative positions of image patches \cite{Doersch2015} may reflect feasible spatial layouts of objects; possible colors can be inferred \cite{Zhang2016,Zhang2017} if the networks can relate colors to object appearances. Rather than rely on temporal changes in video, these methods are able to exploit still images.

Our work is also closely related to mid-level patch clustering \cite{Singh2012,Doersch2013,Doersch2014} and unsupervised discovery of semantic classes \cite{Russell2006,Sivic2005} as we attempt to find reliable clusters in our affinity graph. In addition, the ranking function used in this paper is related to deep metric learning with Siamese architectures \cite{Chopra2005,Hadsell2006,Gong2013,Wang2014,Hoffer2015}.


\vspace{.5em}
\noindent\textbf{Analysis of the two types of invariance.}
Our generic framework can be instantiated by any two self-supervised methods that can respectively learn inter-/intra-instance invariance. In this paper we adopt Doersch \etal's \cite{Doersch2015} context prediction method to build inter-instance invariance, and Wang and Gupta's \cite{Wang2015} tracking method to build intra-instance invariance. We analyze their behaviors as follows.

\begin{figure}
\centering
\includegraphics[width=\linewidth]{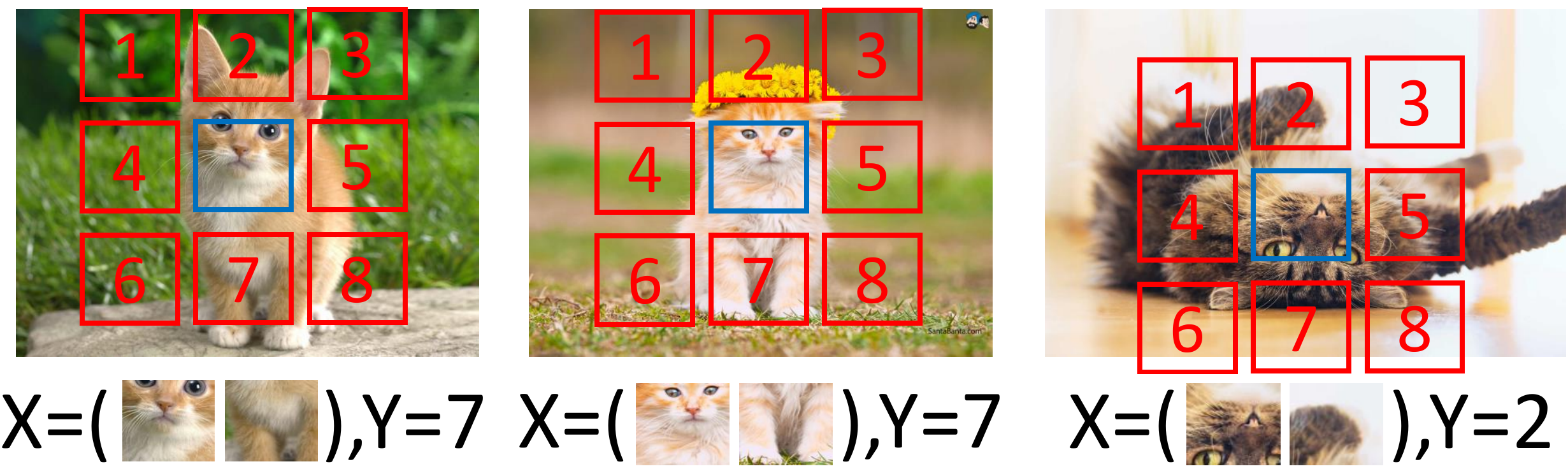}
\vspace{-1.5em}
\caption{The \emph{context prediction} task defined in~\cite{Doersch2015}. Given two patches in an image, it learns to predict the relative position between them.}\label{fig:context}
\vspace{-0.5em}
\end{figure}

The context prediction task in \cite{Doersch2015} randomly samples a patch (blue in Figure~\ref{fig:context}) and one of its eight neighbors (red), and trains the network to predict their relative position, defined as an 8-way classification problem. In the first two examples in Figure~\ref{fig:context}, the context prediction model is able to predict that the ``leg" patch is below the ``face'' patch of the cat, indicating that the model has learned some commonality of spatial layout from the training data. However, the model would fail if the pose, viewpoint, or deformation of the object is changed drastically, \eg, in the third example of Figure~\ref{fig:context} --- unless the dataset is diversified and large enough to include gradually changing poses, it is hard for the models to learn that the changed pose can be of the same object type.

On the other hand, these changes can be more successfully captured by the visual tracking method presented in \cite{Wang2015}, \eg, see \br{A}{A'} and \br{B}{B'} in Figure~\ref{fig:teaser}. But by tracking an identical instance we cannot associate different instances of the same semantics. Thus we expect the representations learned in \cite{Wang2015} are weak in handling the variations between different objects in the same category. 

\section{Overview} 

\begin{figure}
\centering
\includegraphics[width=.8\linewidth]{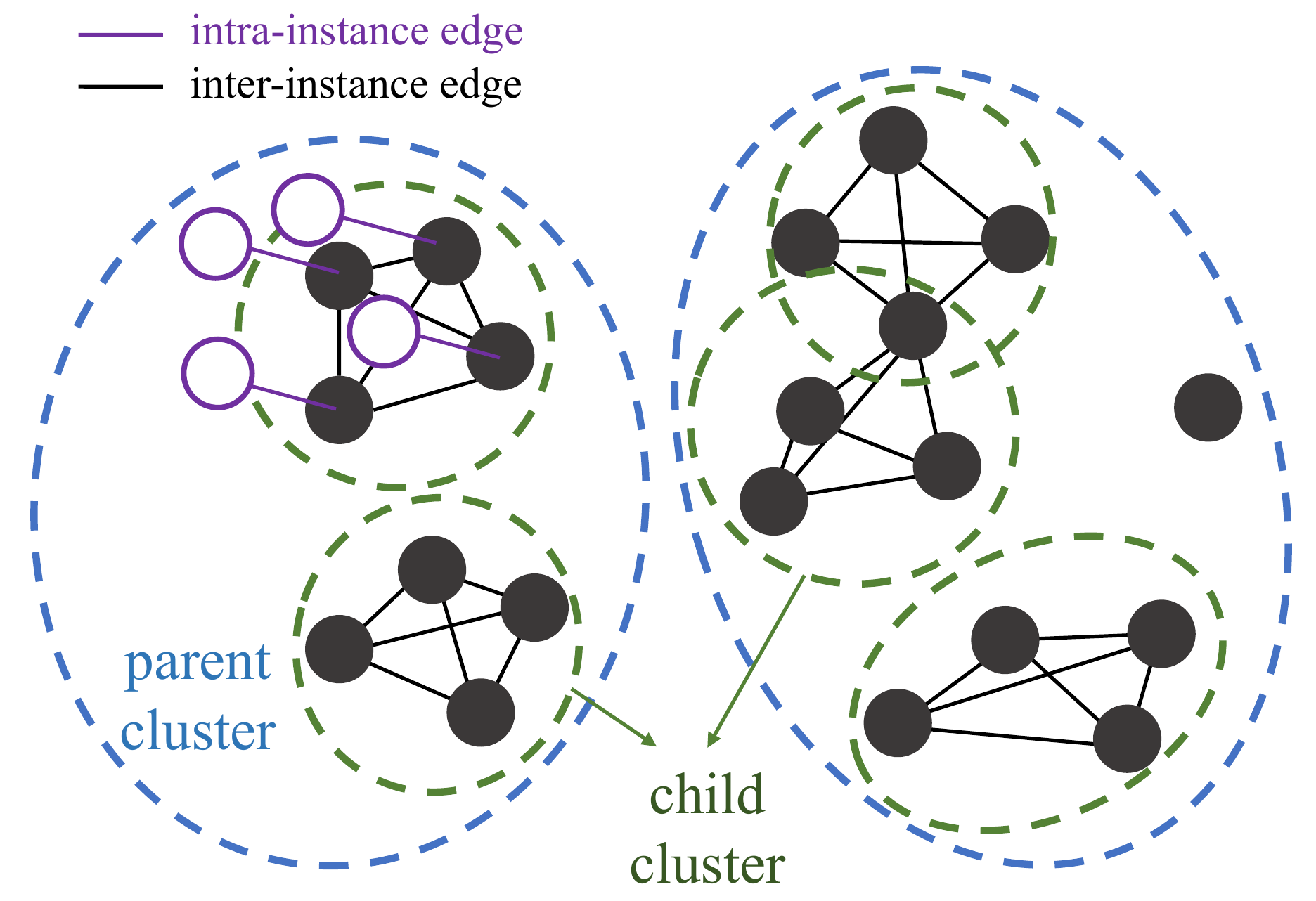}
\vspace{-0.5em}
\caption{Illustrations for our graph construction. We first cluster the object nodes into coarser clusters (namely ``parent'' clusters) and then inside each cluster we perform nearest-neighbor search to obtain ``child'' clusters consisting of $4$ samples. Samples in each child cluster are linked to each other with the ``inter-instance'' edges. We add new samples via visual tracking and link them to the original objects by ``intra-instance'' edges. }\label{fig:graph}
\vspace{-0.5em}
\end{figure}

Our goal is to learn visual representations which capture: (i) inter-instance invariance (\eg, two instances of cats should have similar features), and (ii) intra-instance invariance (pose, viewpoint, deformation, illumination, and other variance of the same object instance). We have tried to formulate this as a multi-task (multi-loss) learning problem in our initial experiments (detailed in Table~\ref{tab:voc07_frcn_ablative} and \ref{tab:voc07_fasterrcnn}) and observed unsatisfactory performance. Instead of doing so, we propose to obtain a richer set of invariance by performing transitive reasoning on the data.

Our first step is to construct a graph that describes the affinity among image patches. A node in the graph denotes an image patch. We define two types of edges in the graph that relate image patches to each other. The first type of edges, called \emph{inter-instance} edges, link two nodes which correspond to different object instances of similar visual appearance; the second type of edges, called \emph{intra-instance} edges, link two nodes which correspond to an identical object captured at different time steps of a track. 
The solid arrows in Figure~\ref{fig:teaser} illustrate these two types of edges.

Given the built graph, we want to \emph{transit} the relations via the known edges and associate unconnected nodes that may provide under-explored invariance (Figure~\ref{fig:teaser}, dash arrows). Specifically, as shown in Figure~\ref{fig:teaser}, if patches \br{A}{B} are linked via an inter-instance edge and \br{A}{A'} and \br{B}{B'} respectively are linked via ``intra-instance'' edges, we hope to enrich the invariance by simple transitivity and relate three new pairs of: \br{A'}{B'}, \br{A}{B'}, and \br{A'}{B} (Figure~\ref{fig:teaser}, dash arrows).

We train a Triplet-Siamese network that encourages similar visual representations between the invariant samples (\eg, any pair consisting of $A, A', B, B'$) and at the same time discourages similar visual representations to a third distractor sample (\eg, a random sample $C$ unconnected to $A, A', B, B'$). In all of our experiments, we apply VGG16 \cite{Simonyan2014} as the backbone architecture for each branch of this Triplet-Siamese network. The visual representations learned by this backbone architecture are evaluated on other recognition tasks.

\section{Graph Construction} 

We construct a graph with inter-instance and intra-instance edges. 
Firstly, we apply the method of \cite{Wang2015} on a large set of 100K unlabeled videos (introduced in \cite{Wang2015}) and mine millions of moving objects using motion cues (Sec.~\ref{sec:mining}). We use the image patches of them to construct the nodes of the graph.

We instantiate inter-instance edges by the self-supervised method of \cite{Doersch2015} that learns context predictions on a large set of still images, which provide features to cluster the nodes and set up inter-instance edges (Sec.~\ref{sec:clustering}). On the other hand, we connect the image patches in the same visual track by intra-instance edges (Sec.~\ref{sec:tracking}).

\subsection{Mining Moving Objects} 
\label{sec:mining}

We follow the approach in \cite{Wang2015} to find the moving objects in videos. As a brief introduction, this method first applies Improved Dense Trajectories (IDT) \cite{Wang2013} on videos to extract SURF \cite{Bay2006} feature points and their motion. The video frames are then pruned if there is too much motion (indicating camera motion) or too little motion (\eg, noisy signals). For the remaining frames, it crop a 227$\times$227 bounding box (from $\sim$600$\times$400 images) which includes the most number of moving points as the foreground object. However, for computational efficiency, in this paper we rescale the image patches to $96 \times 96$ after cropping and we use them as inputs for clustering and training.

\subsection{Inter-instance Edges via Clustering}
\label{sec:clustering}

\begin{figure}
\centering
\includegraphics[width=0.8\linewidth]{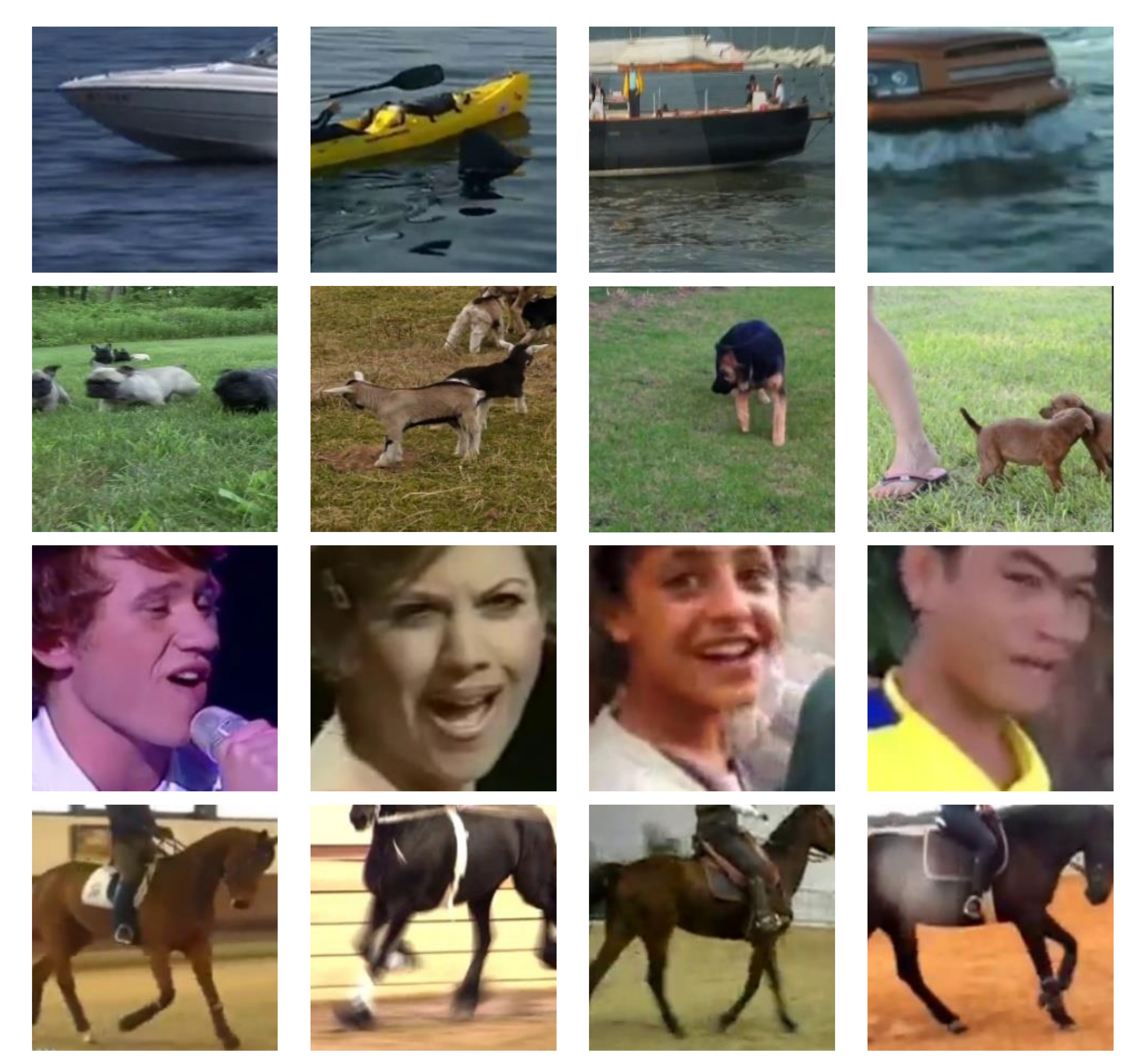}
\vspace{-0.5em}
\caption{Some example clustering results. Each row shows the 4 examples in a child cluster (Sec.~\ref{sec:clustering}).}
\label{fig:clusters}
\end{figure}

Given the extracted image patches which act as nodes, we want to link them with extra inter-instance edges. We rely on the visual representations learned from \cite{Doersch2015} to do this. We connect the nodes representing image patches which are close in the feature space. In addition, motivated by the mid-level clustering approaches~\cite{Singh2012,Doersch2013}, we want to obtain millions of object clusters with a small number of objects in each to maintain high ``purity'' of the clusters. We describe the implementation details of this step as follows.

We extract the \texttt{pool5} features of the VGG16 network trained as in~\cite{Doersch2015}. Following \cite{Doersch2015}, we use ImageNet \emph{without} labels to train this network.
Note that because 
we use a patch size of 96$\times$96, the dimension of our \texttt{pool5} feature is 3$\times$3$\times$512$=$4608. 
The distance between samples is calculated by the cosine distance of these features.
We want the object patches in each cluster to be close to each other in the feature space, and we care less about the differences between clusters.
However, directly clustering  millions of image patches into millions of small clusters (\eg, by K-means) is time consuming. So we apply a hierarchical clustering approach (2-stage in this paper) where we first group the images into a relatively small number of clusters, and then find groups of small number of examples inside each cluster via nearest-neighbor search.

Specifically, in the first stage of clustering, we apply K-means clustering with $K=5000$ on the image patches. We then remove the clusters with number of examples less than $100$ (this reduces $K$ to 546 in our experiments on the image patches mined from the video dataset). We view these clusters as the ``parent'' clusters (blue circles in Figure~\ref{fig:graph}). Then in the second stage of clustering, inside each parent cluster, we perform nearest-neighbor search for each sample and obtain its top 10 nearest neighbors in the feature space. We then find any group of samples with a group size of 4, inside which all the samples are each other's top-10 nearest neighbors. We call these small clusters with $4$ samples ``child'' clusters (green circles in Figure~\ref{fig:graph}). We then link these image patches with each other inside a child cluster via ``inter-instance'' edges.
Note that different child clusters may overlap, \ie, we allow the same sample to appear in different groups. However, in our experiments we find that most samples appear only in one group. We show some results of clustering in Figure~\ref{fig:clusters}. 

\subsection{Intra-instance Edges via Tracking}
\label{sec:tracking}
To obtain rich variations of viewpoint and deformation changes of the same object instance, we apply visual tracking on the mined moving objects in the videos as in \cite{Wang2015}. More specifically, given a moving object in the video, it applies KCF \cite{Henriques2015} to track the object for $N=30$ frames and obtain another sample of the object in the end of the track. Note that the KCF tracker does not require any human supervision. We add these new objects as nodes to the graph and link the two samples in the same track with an intra-instance edge (purple in Figure~\ref{fig:graph}). 

\begin{figure}
\centering
\includegraphics[width=0.45\textwidth]{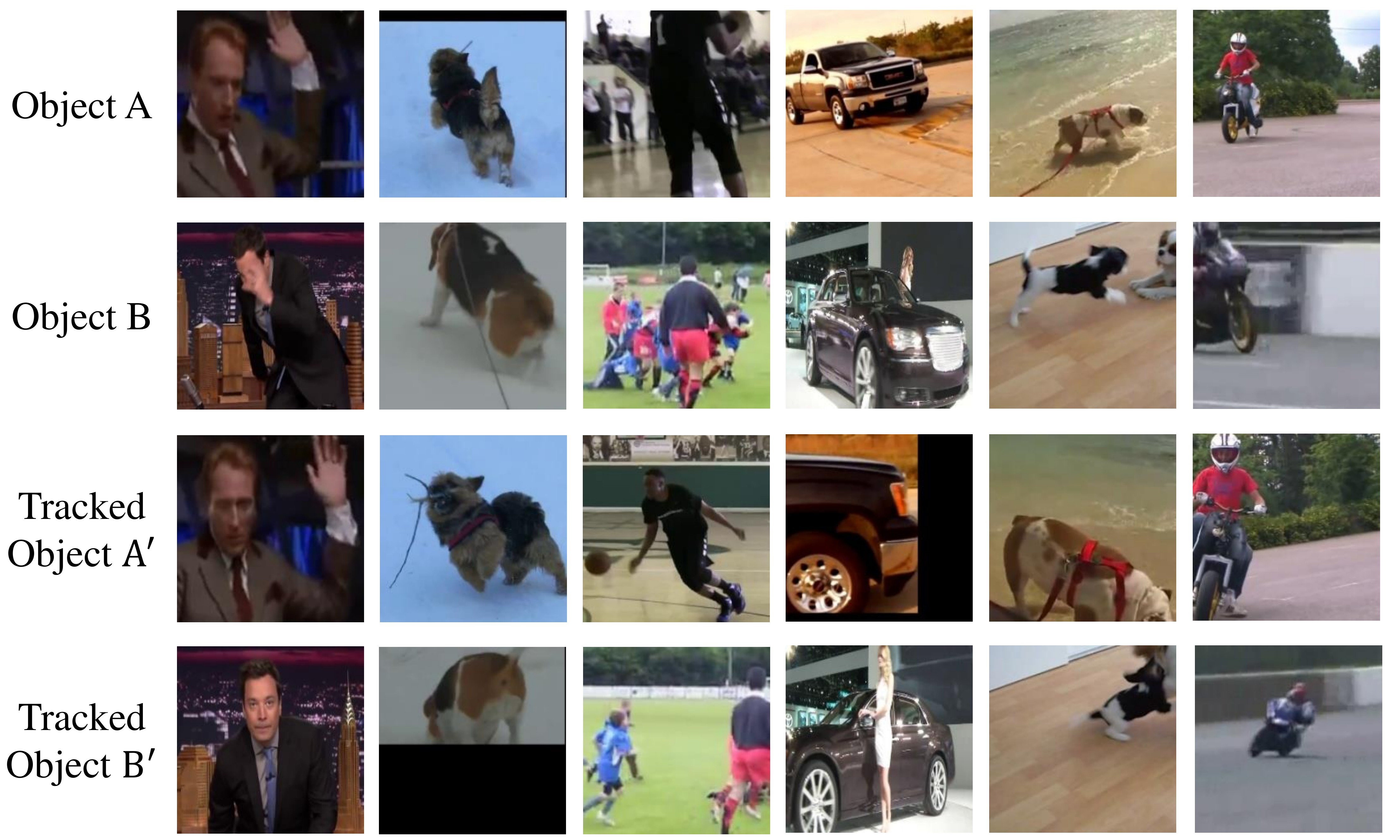}
\caption{Examples used for training the network. Each column shows a set of image patches $\{A, B, A', B'\}$. Here, $A$ and $B$ is linked by an inter-instance edge, and $A'$/$B'$ is linked to $A$/$B$ via intra-instance edges.}\label{fig:examples}
\end{figure}

\begin{figure}
\centering
\includegraphics[width=\linewidth]{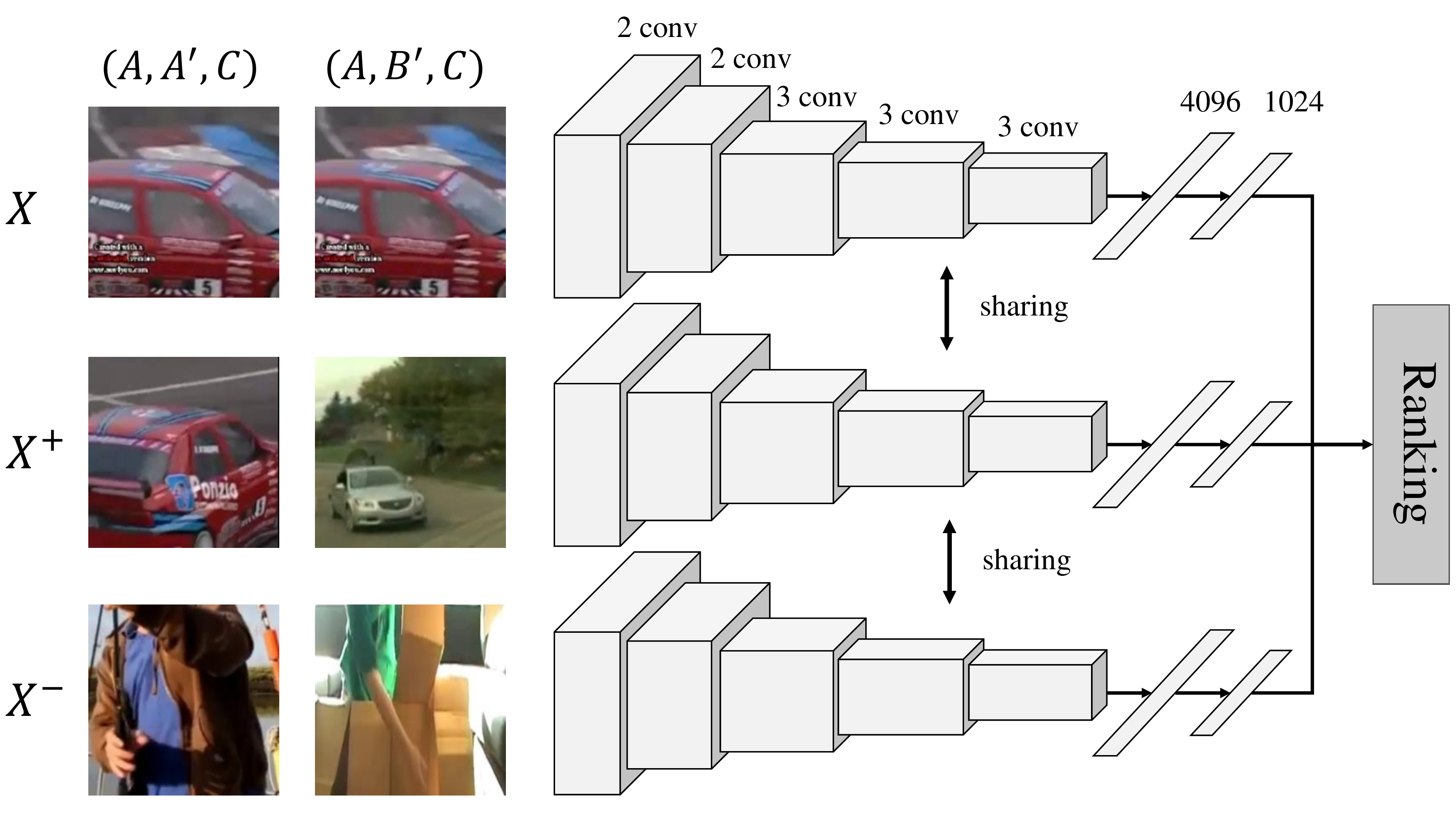}
\caption{Our Triplet-Siamese network. We can feed in the network with different combinations of examples. }\label{fig:network}
\end{figure}

\section{Learning with Transitions in the Graph} 

With the graph constructed, we want to link more image patches (see dotted links in Figure~\ref{fig:teaser}) which may be related via the transitivity of invariance. Objects subject to different levels of invariance can thus be related to each other. Specifically, if we have a set of nodes $\{A,B,A',B'\}$ where \br{A}{B} are connected by an inter-instance edge and \br{A}{A'} and \br{B}{B'} are connected by an intra-instance edge, by assuming transitivity of invariance we expect the new pairs of \br{A}{B'}, \br{A'}{B}, and \br{A'}{B'} to share similar high-level visual representations. Some examples are illustrated in Figure~\ref{fig:teaser} and \ref{fig:examples}.

We train a deep neural network (VGG16) to generates similar visual representations if the image patches are linked by inter-instance/intra-instance edges or their transitivity (which we call a positive pair of samples). To avoid a trivial solution of identical representations, we also encourage the network to generate dissimilar representations if a node is expected to be unrelated. Specifically, we constrain the image patches from different ``parent'' clusters (which are more likely to have different categories) to have different representations (which we call a negative pair of samples). We design a Triplet-Siamese network with a ranking loss function \cite{Wang2014,Wang2015} such that 
the distance between related samples should be smaller than the distance of unrelated samples.

Our Triplet-Siamese network includes three towers of a ConvNet with shared weights (Figure~\ref{fig:network}). For each tower, we adopt the standard VGG16 architecture \cite{Simonyan2014} to the convolutional layers, after which we add two fully-connected layers with $4096$-d and $1024$-d outputs. The Triplet-Siamese network accepts a triplet sample as its input: the first two image patches in the triplet are a positive pair, and the last two are a negative pair. We extract their $1024$-d features and calculate the ranking loss as follows.

Given an arbitrary pair of image patches $A$ and $B$, we define their distance as: $D(A, B) = 1 - \frac{F(A) \cdot F(B)} {\|F(A)\| \|F(B)\| }$ where $F(\cdot)$ is the representation mapping of the network. With a triplet of $(X, X^{+}, X^{-})$ where $(X, X^{+})$ is a positive pair and $(X, X^{-})$ is a negative pair as defined above, we minimize the ranking loss:
\begin{equation}\label{eq:loss}
\mathcal{L}(X, X^{+}, X^{-}) =  \max\{0, \mathcal{D}(X, X^{+}) - \mathcal{D}(X, X^{-}) + m \}, \nonumber
\end{equation}
where $m$ is a margin set as $0.5$ in our experiments.
Although we have only one objective function, we have different types of training examples. As illustrated in Figure \ref{fig:network}, given the set of related samples $\{A,B,A',B'\}$ (see Figure~\ref{fig:examples}) and a random distractor sample $C$ from another parent cluster, we can train the network to handle, \eg, viewpoint invariance for the same instance via $\mathcal{L}(A, A', C)$ and invariance to different objects sharing the same semantics via $\mathcal{L}(A, B', C)$.

Besides exploring these relations, we have also tried to enforce the distance between different objects to be larger than the distance between two different viewpoints of the same object, \eg, $\mathcal{D}(A,A') < \mathcal{D}(A,B')$. But we have not found this extra relation brings any improvement. 
Interestingly, we found that the representations learned by our method can in general satisfy $\mathcal{D}(A,A') < \mathcal{D}(A,B')$ after training.

\section{Experiments}

\begin{figure*}
\centering
\includegraphics[width=1\textwidth]{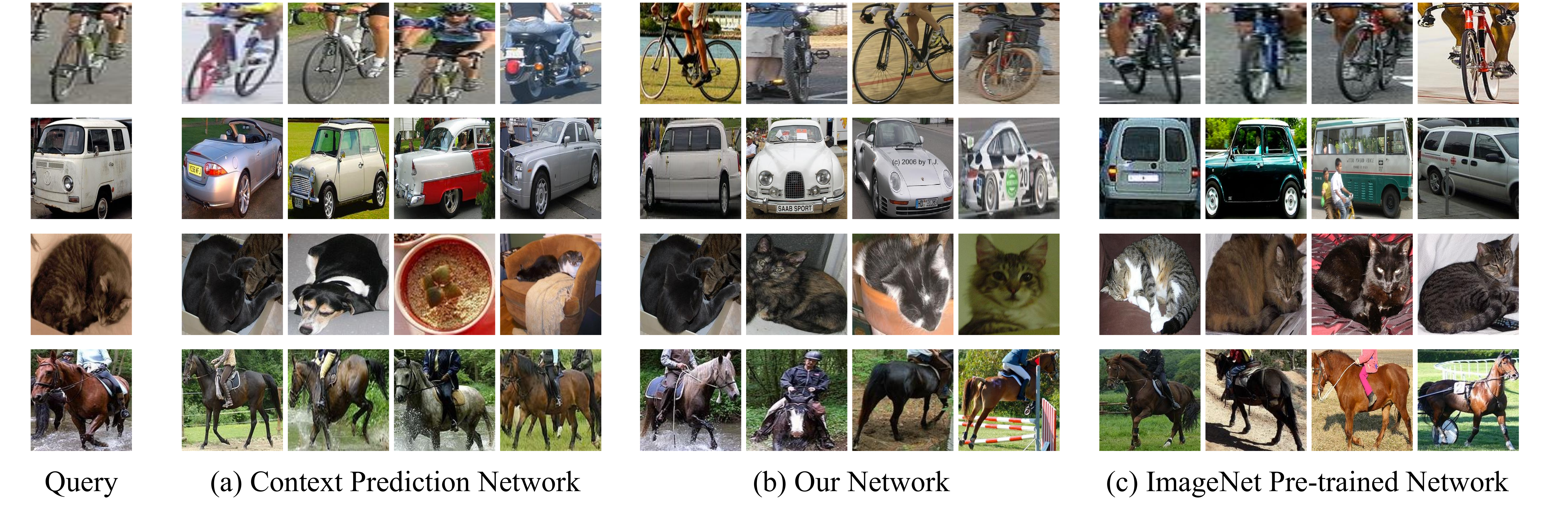}
\vspace{-2em}
\caption{Nearest-neighbor search on the PASCAL VOC dataset. We extract three types of features: (a) context prediction network from~\cite{Doersch2015}, (b) network trained with our self-supervised method, and (c) the network pre-trained in the annotated ImageNet dataset. We show that our network can represent a greater variety (\eg, viewpoints) of objects of the same category.}\label{fig:nn_results}
\end{figure*}

We perform extensive analysis on our self-supervised representations. We first evaluate our ConvNet as a feature extractor on different tasks \emph{without} fine-tuning . We then show the results of transferring the representations to vision tasks including object detection and surface normal estimation with fine-tuning. 

\vspace{.5em}
\noindent\textbf{Implementation Details.} To prepare the data for training, we download the 100K videos from YouTube using the URLs provided by \cite{Liang2014,Wang2015}. By mining the moving objects and tracking in the videos, we obtain $\sim$10 million image patches of objects. By applying the transitivity on the graph constructed, we obtain 7 million positive pairs of objects where each pair of objects are two different instances with different viewpoints. We also randomly sample 2 million object pairs connected by the intra-instance edges. 

We train our network with these 9 million pairs of images using a learning rate of $0.001$ and a mini-batch size of $100$. For each pair we sample the third distractor patch from a different ``parent cluster'' in the same mini-batch. We use the network pre-trained in \cite{Doersch2015} to initialize our convolutional layers and randomly initialized the fully connected layers. We train the network for 200K iterations with our method. 

\subsection{Qualitative Results without Fine-tuning}

We first perform nearest-neighbor search to show qualitative results. We adopt the pool5 feature of the VGG16 network for all methods without any fine-tuning (Figure~\ref{fig:nn_results}). We do this experiment on the object instances cropped from the PASCAL VOC 2007 dataset \cite{Everingham2010} (\texttt{trainval}).  As Figure~\ref{fig:nn_results} shows, given an query image on the left, the network pre-trained with the context prediction task~\cite{Doersch2015} can retrieve objects of very similar viewpoints. On the other hand, our network shows more variations of objects and can often retrieve objects with the same class as the query. We also show the nearest-neighbor results using fully-supervised ImageNet pre-trained features as a comparison. 

\begin{figure}[t]
\centering
\includegraphics[width=0.45\textwidth]{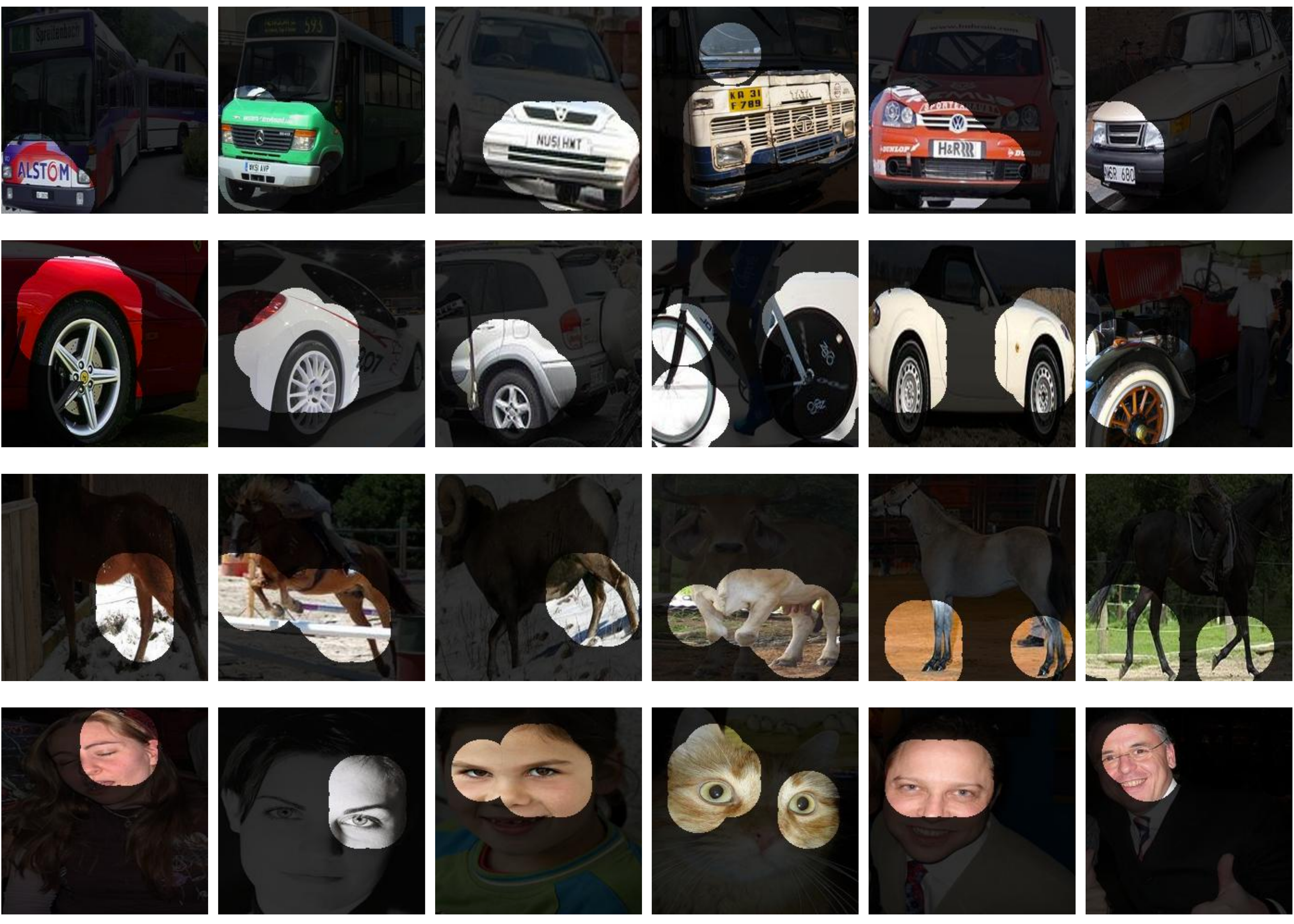}
\vspace{-0.5em}
\caption{Top 6 responses for neurons in 4 different convolutional units of our network, visualized using \cite{Zhou2015}.}\label{fig:RF}
\vspace{-0.5em}
\end{figure}

We also visualize the features using the visualization technique of \cite{Zhou2015}. For each convolutional unit in conv5$\_3$, we retrieve the objects which give highest activation responses and highlight the receptive fields on the images. We visualize the top 6 images for 4 different convolutional units in Figure~\ref{fig:RF}. We can see these convolutional units are corresponding to different semantic object parts (\eg, fronts of cars or buses wheels, animal legs, eyes or faces).  

\begin{table*}[t]
\centering
\renewcommand{\arraystretch}{1.1}
\renewcommand{\tabcolsep}{1.2mm}
\resizebox{\linewidth}{!}{
\begin{tabular}{@{}L{2.1cm} !{\color{gray}\vrule}  c !{\color{gray}\vrule} c*{19}{x} @{}}
\Xhline{1pt}
method   & mAP & aero      & bike      & bird      & boat      & bottle     & bus        & car        & cat        & chair      & cow        & table      & dog        & horse      & mbike      & persn     & plant      & sheep      & sofa       & train      & tv   \\
\Xhline{0.8pt}
from scratch    & 39.7 & 51.7 & 55.8 & 21.7 & 24.0 & 10.5 & 58.7 & 59.2 & 41.1 & 18.2 & 32.9 & 35.6 & 33.4 & 60.4 & 57.3 & 45.5 & 19.7 & 29.2 & 30.8 & 61.0 & 47.3  \\
Vid-Edge~\cite{Li2016}    & 44.2 & 54.4 & 58.2 & 39.6 & 30.8 & 12.5 & 58.7 & 61.9 & 51.0 & 22.0 & 41.4 & 47.4 & 41.5 & 63.2 & 58.4 & 47.5 & 17.2 & 27.6 & 45.4 & 59.8 & 45.4   \\
Context~\cite{Doersch2015}    & 61.5 & 70.8 & 72.1 & 54.7 & 49.7 & 31.0 & 72.3 & 76.9 & 70.8 & 44.6 & 61.1 & 59.8 & 67.0 & 74.6 & 72.5 & 68.3 & 29.4 & 58.5 & 66.9 & 75.1 & 54.3   \\
Tracking~\cite{Wang2015} & 60.2  & 65.7 & 73.2 & 55.4 & 46.4 & 30.9 & 74.0 & 76.9 & 67.8 & 40.9 & 58.0 & 60.9 & 65.0 & 74.1 & 71.6 & 67.1 & 31.5 & 55.0 & 61.8 & 73.9 & 53.8 \\
Ours  & 63.2 & 68.4 & 74.6 & 57.1 & 49.6 & 34.1 & 73.5 & 76.9 & 73.2 & 45.8 & 63.3 & 66.3 & 68.6 & 74.9 & 74.2 & 69.5 & 31.9 & 57.4 & 70.3 & 75.9 & 59.3  \\
\hline
ImageNet   & 67.3 & 74.4 & 78.0 & 65.9 & 54.4 & 39.7 & 76.4 & 78.6 & 82.5 & 48.6 & 73.3 & 67.2 & 78.4 & 77.3 & 75.7 & 72.2 & 32.2 & 65.8 & 66.8 & 75.2 & 62.4 \\
\Xhline{1pt}
\end{tabular}
}
\vspace{-.5em}
\caption{\small Object detection Average Precision (\%) on the VOC 2007 test set using \textbf{Fast} R-CNN \cite{Girshick2015} (with selective search proposals \cite{Uijlings2013}): comparisons among different self-supervised learning approaches.}
\label{tab:voc07_frcn}
\end{table*}

\begin{table*}[t]
\centering
\renewcommand{\arraystretch}{1.1}
\renewcommand{\tabcolsep}{1.2mm}
\resizebox{\linewidth}{!}{
\begin{tabular}{@{}L{2.4cm} !{\color{gray}\vrule}  c !{\color{gray}\vrule} c*{19}{x} @{}}
\Xhline{1pt}
method   & mAP & aero      & bike      & bird      & boat      & bottle     & bus        & car        & cat        & chair      & cow        & table      & dog        & horse      & mbike      & persn     & plant      & sheep      & sofa       & train      & tv   \\
\Xhline{0.8pt}
Ours  & 63.2 & 68.4 & 74.6 & 57.1 & 49.6 & 34.1 & 73.5 & 76.9 & 73.2 & 45.8 & 63.3 & 66.3 & 68.6 & 74.9 & 74.2 & 69.5 & 31.9 & 57.4 & 70.3 & 75.9 & 59.3  \\
Multi-Task & 62.1  & 70.0 & 74.2 & 57.2 & 48.4 & 33.0 & 73.6 & 77.6 & 70.7 & 45.0 & 61.5 & 64.8 & 67.2 & 74.0 & 72.9 & 68.3 & 32.4 & 56.6 & 64.1 & 74.1 & 57.5  \\
Ours (15-frame)    & 61.5 &  70.3 & 74.1 & 53.3 & 47.1 & 33.5 & 74.6 & 77.1 & 67.7 & 43.3 & 58.1 & 65.5 & 65.8 & 75.2 & 72.2 & 67.6 & 31.6 & 55.5 & 65.6 & 74.6 & 57.2   \\
Ours (HOG)    & 60.4 & 65.8 & 73.4 & 54.7 & 47.7 & 30.2 & 75.6 & 77.1 & 67.6 & 42.0 & 58.8 & 63.2 & 65.3 & 74.1 & 72.0 & 67.2 & 29.9 & 54.4 & 62.1 & 72.9 & 53.9  \\
\Xhline{1pt}
\end{tabular}
}
\vspace{-0.5em}
\caption{\small More ablative studies on object detection on the VOC 2007 test set using \textbf{Fast} R-CNN \cite{Girshick2015} (with selective search proposals \cite{Uijlings2013}).}
\label{tab:voc07_frcn_ablative}
\end{table*}

\subsection{Analysis on Object Detection}

We evaluate how well our representations can be transferred to object detection by fine-tuning Fast R-CNN \cite{Girshick2015} on PASCAL VOC 2007 \cite{Everingham2010}. We use the standard \texttt{trainval} set for training and \texttt{test} set for testing with VGG16 as the base architecture. For the detection network, we initialize the weights of convolutional layers from our self-supervised network and randomly initialize the fully-connected layers using Gaussian noise with zero mean and $0.001$ standard deviation. 

During fine-tuning Fast R-CNN, we use $0.00025$ as the starting learning rate. We reduce the learning rate by 1/10 in every 50K iterations. We fine-tune the network for 150K iterations. Unlike standard Fast R-CNN where the first few convolutional layers of the ImageNet pre-trained network are fixed, we fine-tuned all layers on the PASCAL data as our model is pre-trained in a very different domain (\eg, video patches).

We report the results in Table~\ref{tab:voc07_frcn}. If we train Fast R-CNN \emph{from scratch} without any pre-training, we can only obtain $39.7\%$ mAP. With our self-supervised trained network as initialization, the detection mAP is increased to $63.2\%$ (with a $23.5$ points improvement). Our result compares competitively (4.1 points lower) to the counterpart using ImageNet pre-training ($67.3\%$ with VGG16).

As we incorporate the invariance captured from \cite{Wang2015} and \cite{Doersch2015}, we also evaluate the results using these two approaches individually (Table~\ref{tab:voc07_frcn}). By fine-tuning the context prediction network of \cite{Doersch2015}, we can obtain $61.5\%$ mAP. To train the network of \cite{Wang2015}, we use exactly the same loss function and initialization as our approach except that there are only training examples of the same instance in the same visual track (\ie, only the samples linked by intra-instance edges in our graph). Its results is $60.2\%$ mAP. 
Our result ($63.2\%$) is better than both methods. This comparison indicates the effectiveness of exploiting a greater variety of invariance in representation learning.

\vspace{.5em}
\noindent\textbf{Is multi-task learning sufficient?} An alternative way of obtaining both intra- and inter-instance invariance is to apply multi-task learning with the two losses of \cite{Doersch2015} and \cite{Wang2015}. Next we compare with this method.

For the task in~\cite{Wang2015}, we use the same network architecture as our approach; for the task in~\cite{Doersch2015}, we follow their design of a Siamese network. We apply different fully connected layers for different tasks, but share the convolutional layers between these two tasks. Given a mini-batch of training samples, we perform ranking among these images as well as context prediction in each image simultaneously via two losses. The representations learned in this way, when fine-tuned with Fast R-CNN, obtain $62.1\%$ mAP (``Multi-task'' in Table~\ref{tab:voc07_frcn_ablative}). Comparing to only using context prediction \cite{Doersch2015} ($61.5\%$), the multi-task learning only gives a marginal improvement ($0.6\%$). This result suggests that multi-task learning in this way is not sufficient; organizing and exploiting the relationships of data, as done by our method, is more effective for representation learning.

\vspace{.5em}
\textbf{How important is tracking?} To further understand how much visual tracking helps, we perform ablative analysis by making the visual tracks shorter: we track the moving objects for $15$ frames instead of  by default $30$ frames. This is expected to reduce the viewpoint/pose/deformation variance contributed by tracking. Our model pre-trained in this way shows $61.5\%$ mAP (``15-frame'' in Table~\ref{tab:voc07_frcn_ablative}) when fine-tuned for detection.
This number is similar to that of using context prediction only (Table~\ref{tab:voc07_frcn}). This result is not surprising, because it does not add much new information for training. It suggests adding stronger viewpoint/pose/deformation invariance is important for learning better features for object detection. 

\vspace{.5em}
\textbf{How important is clustering?} Furthermore, we want to understand how important it is to cluster images with features learned from still images \cite{Doersch2015}.  We perform another ablative analysis by replacing the features of \cite{Doersch2015} with HOG \cite{Dalal2005} during clustering. The rest of the pipeline remains exactly the same. The final result is $60.4\%$ mAP (``HOG'' in Table~\ref{tab:voc07_frcn_ablative}). This shows that if the features for clustering are not invariant enough to handle different object instances, the transitivity in the graph becomes less reliable. 

\subsection{Object Detection with Faster R-CNN}

Although Fast R-CNN \cite{Girshick2015} has been a popular testbed of un-/self-supervised features, it relies on Selective Search proposals \cite{Uijlings2013} and thus is not fully end-to-end.
We further evaluate the representations on object detection with the \emph{end-to-end} Faster R-CNN~\cite{Ren2015} where the Region Proposal Network (RPN) may suffer from the features if they are low-quality.

\begin{table}
\renewcommand{\arraystretch}{1.0}
\centering
\small
\begin{tabular}{@{}l@{ }c@{ }c@{ }c@{ }c@{ }c@{ }c@{ }c} \toprule
        & All~~~ & $>$c1~~~ & $>$c2~~~ & $>$c3~~~ & $>$c4~~~ & $>$c5~~~ \\ \midrule
Context~\cite{Doersch2015}   & 62.6  &  61.1  & 60.9  & 57.0 & 49.7 & 38.1 \\
Tracking~\cite{Wang2015}   & 62.2  &  61.5  & 62.2  & 61.4 & 58.9 & 39.5 \\
Multi-Task~\cite{Doersch2015,Wang2015} & 62.4  &  63.2  & 63.5  & 62.9 & 58.7 & 27.6 \\
Ours      & 65.0 &  64.5  & 63.6  & 60.4 & 55.7 & 43.1 \\ \midrule
ImageNet  & 70.9 &  71.1  & 71.1  & 70.2 & 70.3 & 64.3 \\ 
\bottomrule
\end{tabular}
\vspace{-0.5em}
\caption{\small {Object detection Average Precision (\%) on the VOC 2007 test set using joint training \textbf{Faster} R-CNN \cite{Ren2015}.}}
\label{tab:voc07_fasterrcnn}
\end{table}

\begin{table}
\renewcommand{\arraystretch}{1.0}
\centering
\small
\begin{tabular}{@{}l@{ }c@{ }c@{ }c@{ }c@{ }c@{ }c@{ }c} \toprule
        & AP~~ & AP$^{50}$~~ & AP$^{75}$~~ & AP$^{S}$~~ & AP$^{M}$~~ & AP$^{L}$~~ \\ \midrule
from scratch   & 20.5  &  40.1  & 19.0  & 5.6 & 22.5 & 32.7 \\
Context~\cite{Doersch2015}   & 22.7  &  43.5  & 21.2  & 6.6 & 24.9 & 36.5 \\
Tracking~\cite{Wang2015}   & 22.6  &  42.8  & 21.6  & 6.3 & 25.0 & 36.2 \\
Multi-Task~\cite{Doersch2015,Wang2015} & 22.0  &  42.3  & 21.1  & 6.6 & 24.5 & 35.0 \\
Ours      & 23.5 &  44.4  & 22.6  & 7.1 & 25.9 & 37.3 \\ \midrule
ImageNet (shorter) & 23.7 & 44.5  & 23.5  & 7.2 & 26.9 & 37.4 \\ 
ImageNet & 24.4 &  46.4  & 23.1  & 7.9 & 27.4 & 38.1 \\ 
\bottomrule
\end{tabular}
\vspace{-0.5em}
\caption{\small Object detection Average Precision (\%, COCO definitions) on COCO minival using joint training \textbf{Faster} R-CNN \cite{Ren2015}. ``(shorter)'' indicates a shorter training time (fewer iterations, 61.25K) used by the codebase of \cite{Ren2015}.
}
\label{tab:COCO}
\end{table}

\vspace{.5em}
\textbf{PASCAL VOC 2007 Results.} We fine-tune Faster R-CNN in 8 GPUs for 35K iterations with an initial learning rate of $0.00025$ which is reduced by 1/10 after every 15K iterations.
Table~\ref{tab:voc07_fasterrcnn} shows the results of fine-tuning all layers (``All'') and also ablative results on freezing different levels of convolutional layers (\eg, the column $>$c3 represents freezing all the layers below and including conv3$\_$x in VGG16 during fine-tuning). Our method gets even better results of $65.0\%$ by using Faster R-CNN, showing a larger gap compared to the counterparts of \cite{Doersch2015} ($62.6\%$) and \cite{Wang2015} ($62.2\%$). 
Noteworthily, when freezing all the convolutional layers and only fine-tuning the fully-connected layers, our method ($43.1\%$) is much better than other competitors. And we again find that the multi-task alternative does not work well for Faster R-CNN.

\vspace{.5em}
\textbf{COCO Results.}  We further report results on the challenging COCO detection dataset \cite{Lin2014}. To the best of our knowledge this is the first work of this kind presented on COCO detection. We fine-tune Faster R-CNN in 8 GPUs for 120K iterations with an initial learning rate of $0.001$ which is reduced by 1/10 after 80k iterations. This is trained on the COCO \texttt{trainval35k} split and evaluated on the \texttt{minival5k} split, introduced by \cite{Bell2015}.

We report the COCO results on Table~\ref{tab:COCO}. Faster R-CNN fine-tuned with our self-supervised network obtains  $23.5\%$ AP using the COCO metric, which is very close ($<$$1\%$) to fine-tuning Faster R-CNN with the ImageNet pre-trained counterpart ($24.4\%$). Actually, if the fine-tuning of the ImageNet counterpart follows the ``shorter" schedule in the public code (61.25K iterations in 8 GPUs, converted from 490K in 1 GPU)\footnote{\url{https://github.com/rbgirshick/py-faster-rcnn}}, the ImageNet supervised pre-training version has $23.7\%$ AP and is comparable with ours. This comparison also strengthens the significance of our result.

To the best of our knowledge, our model achieves the best performance reported to date on VOC 2007 and COCO using un-/self-supervised pre-training.

\subsection{Adapting to Surface Normal Estimation}

To show the generalization ability of our self-supervised representations, we adopt the learned network to the surface normal estimation task. In this task, given a single RGB image as input, we train the network to predict the normal/orientation of the pixels. We evaluate our method on the NYUv2 RGBD dataset~\cite{Silberman2012} dataset. We use the official split of 795 images for training and 654 images for testing. We follow the same protocols for generating surface normal ground truth and evaluations as~\cite{Fouhey13a,Ladicky14b,Fouhey14c}. 

To train the network for surface normal estimation, we apply the Fully Convolutional Network (FCN 32-s) proposed in~\cite{Long2015} with the VGG16 network as base architecture. For the loss function, we follow the design in~\cite{Wang2015a}. Specifically, instead of direct regression to obtain the normal, we use a codebook of $40$ codewords to encode the 3-dimension normals. Each codeword represents one class thus we turn the problem into a 40-class classification for each pixel. We use the same hyperparameters as in \cite{Long2015} for training and the network is fine-tuned for same number of iterations (100K) for different initializations.  

To initialize the FCN model with self-supervised nets, we copy the weights of the convolutional layers to the corresponding layers in FCN. For ImageNet pre-trained network, we follow~\cite{Long2015} by converting the fully connected layers to convolutional layers and copy all the weights. For the model trained from scratch, we randomly initialize all the layers with ``Xavier'' initialization \cite{Glorot2010} . 

Table~\ref{tab:3dresults} shows the results. We report mean and median error for all visible pixels (in degrees) and also the percentage of pixels with error less than 11.25, 22.5 and 30 degrees. Surprisingly, we obtain much better results with our self-supervised trained network than ImageNet pre-training in this task ($3$ to $4\%$ better in most metrics). As a comparison, the network trained in~\cite{Doersch2015,Wang2015} are slightly worse than the ImageNet pre-trained network. These results suggest that our learned representations are competitive to ImageNet pre-training for high-level semantic tasks, but outperforms it on tasks such as surface normal estimation. This experiment suggests that different visual tasks may prefer different levels of visual invariance.

\begin{table}
\renewcommand{\arraystretch}{1.0}
\small
\centering
\begin{tabular}{@{}l@{ }c@{ }c@{ }c@{ }c@{ }c@{ }c} \toprule
        & ~Mean~~~ & Median & $11.25^{\circ}$ & $22.5^{\circ}$ & $30^{\circ}$ \\
        & \multicolumn{2}{c}{(lower is better)} & \multicolumn{3}{c}{(higher is better)} \\\midrule
from scratch                 & 31.3      &  25.3  & 24.2  & 45.6      & 56.8 \\
Context~\cite{Doersch2015}               & 29.0      &  21.6  & 28.8  & 51.5      & 61.9 \\
Tracking~\cite{Wang2015}               & 27.8      &  21.8  & 27.4  & 51.1      & 62.5 \\
Ours                & \textbf{26.0}      &  \textbf{18.0}  & \textbf{33.9}  & \textbf{57.6}      & \textbf{67.5} \\
\hline
ImageNet              & 27.8      &  21.2  & 29.0  & 52.3      & 63.4 \\ 
\bottomrule
\end{tabular}
\vspace{-0.5em}
\caption{Results on NYU v2 for per-pixel surface normal estimation, evaluated over valid pixels.}
\label{tab:3dresults}
\end{table}

\vspace{1em}
{\noindent {\bf Acknowledgement}: This work was supported by ONR MURI N000141612007 and Sloan Fellowship to AG.}

{\small
\bibliographystyle{ieee}
\bibliography{local}
}

\end{document}